\documentclass[journal]{IEEEtai}

\usepackage[colorlinks,urlcolor=blue,linkcolor=blue,citecolor=blue]{hyperref}

\usepackage{color,array}

\usepackage{graphicx}

\usepackage{longtable}
\usepackage{booktabs}
\usepackage{float}
\usepackage{subcaption}

\begin{document}

\title{Interpretable Machine Learning for COVID-19: An Empirical Study on Severity Prediction Task}

\author{Han Wu, Wenjie Ruan, Jiangtao Wang, Dingchang Zheng, Bei Liu, Yayuan Geng, Xiangfei Chai, \\
Jian Chen, Kunwei Li, Shaolin Li and Sumi Helal, \IEEEmembership{Fellow, IEEE}
\thanks{This work was supported in part by HY Medical Technology, Scientific Research Department Beijing, CN}
\thanks{Han Wu is with the Coventry University, Priory St, Coventry CV1 5FB UK. (e-mail: hw@exeter.ac.uk).}
\thanks{Wenjie Ruan is with the University of Exeter, Stocker Rd, Exeter EX4 4PY UK (e-mail: W.Ruan@exeter.ac.uk).}
\thanks{Jiangtao Wang is with Coventry University, Priory St, Coventry CV1 5FB UK (e-mail: jiangtao.wang@coventry.ac.uk).}
\thanks{Dingchang Zheng is with Coventry University, Priory St, Coventry CV1 5FB UK (e-mail: ad4291@coventry.ac.uk).}
\thanks{Bei Liu is with the 910 Hospital of PLA, Department of Gastroenterology, CN (e-mail: liubei0927@outlook.com).}
\thanks{Yayuan Geng is with HY Medical Technology, Scientific Research Department
Beijing, CN (e-mail: gengyayuan@huiyihuiying.com).}
\thanks{Xiangfei Chai is with HY Medical Technology, Scientific Research Department
Beijing, CN (e-mail: chaixiangfei@huiyihuiying.com).}
\thanks{Jian Chen is with Fifth Affiliated Hospital of Sun Yat-sen University, Department of Radiology Zhuhai, CN (e-mail: drchenj@126.com).}
\thanks{Kunwei Li is with Fifth Affiliated Hospital of Sun Yat-sen University, Department of Radiology Zhuhai, CN (e-mail: likunwei@mail.sysu.edu.cn).}
\thanks{Shaolin Li is with Fifth Affiliated Hospital of Sun Yat-sen University, Department of Radiology, Zhuhai, CN, and Guangdong Provincial Key Laboratory of Biomedical Imaging
Zhuhai, Guangdong, CN (e-mail: lishlin5@mail.sysu.edu.cn).}
\thanks{Sumi Helal is with University of Florida, Gainesville, Florida, USA (e-mail: helal@acm.org).}}

\markboth{Journal of IEEE Transactions on Artificial Intelligence, Vol. 00, No. 0, Month 2020}{Han. Wu \MakeLowercase{\textit{et al.}}: Interpretable Machine Learning for COVID-19: An Empirical Study on Severity Prediction Task}

\maketitle

\begin{abstract}

The black-box nature of machine learning models hinders the deployment of some high-accuracy medical diagnosis algorithms. It is risky to put one's life in the hands of models that medical researchers do not fully understand or trust. However, through model interpretation, black-box models can promptly reveal significant biomarkers that medical practitioners may have overlooked due to the surge of infected patients in the COVID-19 pandemic.

This research leverages a database of 92 patients with confirmed SARS-CoV-2 laboratory tests between 18th January 2020 and 5th March 2020, in Zhuhai, China, to identify biomarkers indicative of infection severity prediction. Through the interpretation of four machine learning models, decision tree, random forests, gradient boosted trees, and neural networks using permutation feature importance, Partial Dependence Plot (PDP), Individual Conditional Expectation (ICE), Accumulated Local Effects (ALE), Local Interpretable Model-agnostic Explanations (LIME), and Shapley Additive Explanation (SHAP), we identify an increase in N-Terminal pro-Brain Natriuretic Peptide (NTproBNP), C-Reaction Protein (CRP), and lactic dehydrogenase (LDH), a decrease in lymphocyte (LYM) is associated with severe infection and an increased risk of death, which is consistent with recent medical research on COVID-19 and other research using dedicated models. We further validate our methods on a large open dataset with 5644 confirmed patients from the Hospital Israelita Albert Einstein, at São Paulo, Brazil from Kaggle, and unveil leukocytes, eosinophils, and platelets as three indicative biomarkers for COVID-19.

\end{abstract}

\begin{IEEEImpStatement}
The pandemic is a race against time. We seek to answer the question, how can medical practitioners employ machine learning to win the race in the pandemic? Instead of targeting at a high-accuracy black-box model that is difficult to trust and deploy, we use model interpretation that incorporates medical practitioners' prior knowledge to promptly reveal the most important indicators in early diagnosis, and thus win the race in the pandemic.
\end{IEEEImpStatement}


\section{Introduction}


\begin{figure*}
\centering
\includegraphics[width=6.0in]{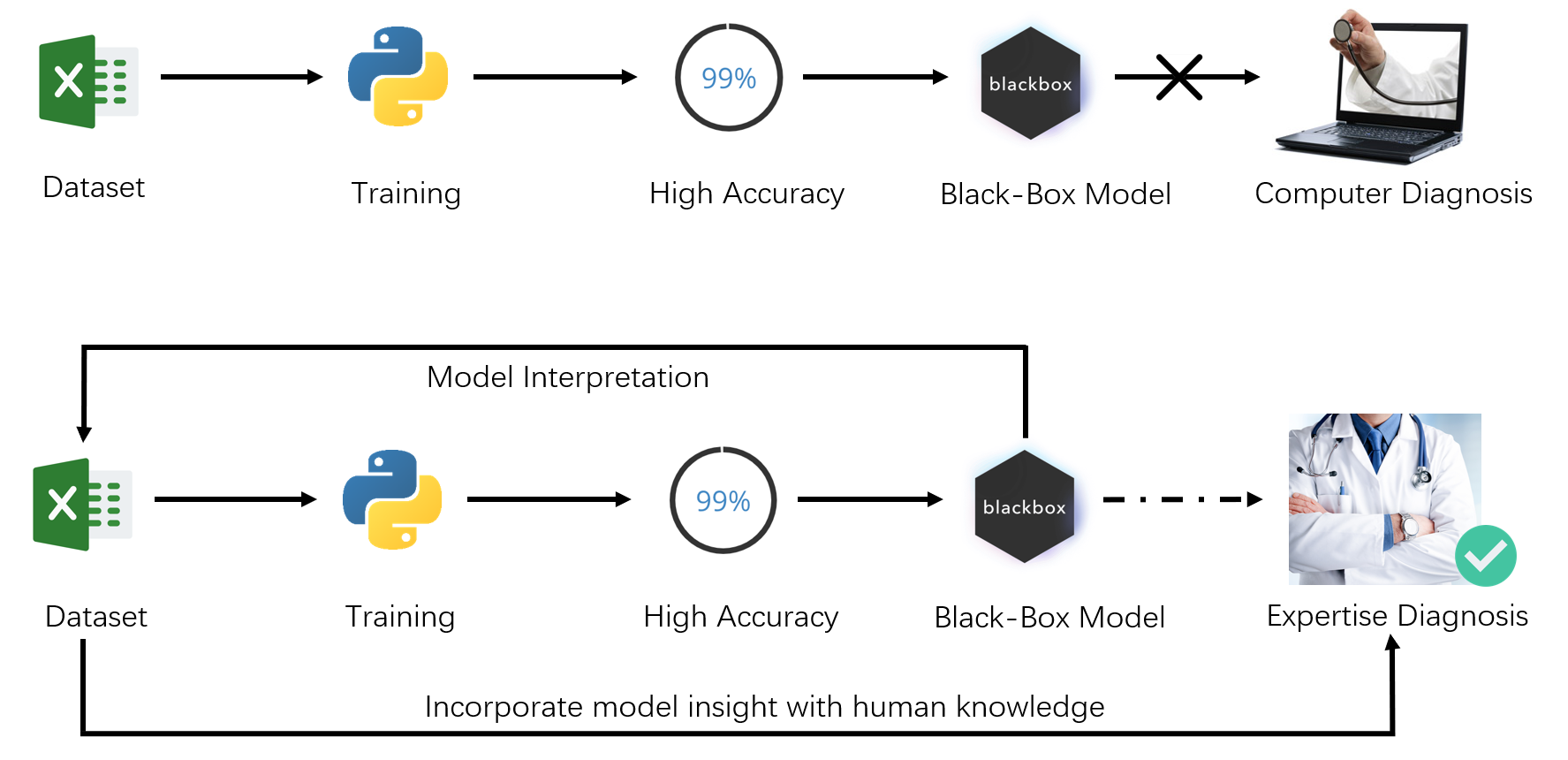}
\caption{The difference between the usual workflow of machine learning, and our approach.}
\label{fig:approach}
\end{figure*}

\IEEEPARstart{T}he sudden outbreak of COVID-19 has caused an unprecedented disruption and impact worldwide. With more than 100 million confirmed cases as of February 2021, the pandemic is still accelerating globally. The disease is transmitted by inhalation or contact with infected droplets with an incubation period ranging from 2 to 14 days \cite{singhal2020review}, making it highly infectious and difficult to contain and mitigate.

With the rapid transmission of COVID-19, the demand for medical supplies goes beyond hospitals' capacity in many countries. Various diagnostic and predictive models are employed to release the pressure on healthcare workers. For instance, a deep learning model that detects abnormalities and extract key features of the altered lung parenchyma using chest CT images is proposed \cite{Basu2020Deep}. On the other hand, Rich Caruana et al. exploit intelligible models that use generalized additive models with pairwise interactions to predict the probability of readmission \cite{Caruana2015IntelligibleMF}. To maintain both interpretability and complexity, DeepCOVIDNet is present to achieve predictive surveillance that identifies the most influential features for the prediction of the growth of the pandemic\cite{ramch2020deepcovidnet} through the combination of two modules. The embedding module takes various heterogeneous feature groups as input and outputs an equidimensional embedding corresponding to each feature group. The DeepFM \cite{guo2017deepfm} module computes second and higher-order interactions between them.



Models that achieves high accuracy provide fewer interpretations due to the trade-off between accuracy and interpretability \cite{doshivelez2017rigorous}. To be adopted in healthcare systems that require both interpretability and robustness\cite{goodfellow2014explaining}, the Multi-tree XGBoost algorithm is employed to identify the most significant indicators in COVID-19 diagnosis \cite{Yan2020}. This method exploits the recursive tree-based decision system of the model to achieve high interpretability. On the other hand, a more complex convolutional neural network (CNN) model can discriminates COVID-19 from Non-COVID-19 using chest CT image \cite{matsuyama2020deep}. It achieves interpretability through gradient-weighted class activation mapping to produce a heat map that visually verifies where the CNN model is focusing.

Besides, several model-agnostic methods have been proposed to peek into black-box models, such as Partial Dependence Plot (PDP) \cite{Friedman2001GreedyFA}, Individual Conditional Expectation (ICE) \cite{goldstein2013peeking}, Accumulated Local Effects (ALE) \cite{apley2016visualizing}, Permutation Feature Importance \cite{fisher2018models}, Local Interpretable Model-agnostic Explanations (LIME) \cite{ribeiro2016i}, Shapley Additive Explanation (SHAP) \cite{lundberg2017unified}, and Anchors \cite{Ribeiro2018AnchorsHM}. Most of these model-agnostic methods are reasoned qualitatively through illustrative figures and human experiences. To quantitatively measure their interpretability, metrics such as faithfulness \cite{alvarezmelis2018robust} and monotonicity \cite{luss2019generating} are proposed.

In this paper, instead of targeting a high-accuracy model, we interpret several models to help medical practitioners promptly discover the most significant biomarkers in the pandemic.

Overall, this paper makes the following contributions:

\begin{enumerate}
    \item \textbf{Evaluation}: A systematic evaluation of the interpretability of machine learning models that predict the severity level of COVID-19 patients. We experiment with six interpretation methods and two evaluation metrics on our dataset and receive the same result as research that uses a dedicated model. We further validate our approach on a dataset from Kaggle.

    \item \textbf{Implication}: Through the interpretation of models trained on our dataset, we reveal N-Terminal pro-Brain Natriuretic Peptide (NTproBNP), C-Reaction Protein (CRP), lactic dehydrogenase (LDH), and lymphocyte (LYM) as the most indicative biomarkers in identifying patients' severity level. Applying the same approach on the Kaggle dataset, we further unveil three significant features, leukocytes, eosinophils, and platelets.

    \item \textbf{Implementation}: We design a system that healthcare professionals can interact with its AI Models to incorporate model insights with medical knowledge. We release our implementation, models for future research and validation. \footnote{Our source code and models are available at \url{https://github.com/wuhanstudio/interpretable-ml-covid-19}.}
\end{enumerate}

\newpage
\section{Preliminary of AI Interpretability}

In this section, six interpretation methods, Partial Dependence Plot, Individual Conditional Expectation, Accumulated Local Effects, Local Interpretable Model-agnostic Explanations, and Shapley Additive Explanation are summarized. We also summarize two evaluation metrics, faithfulness and monotonicity.

\subsection{\textbf{Model-Agnostic Methods}}

In healthcare, restrictions to using only interpretable models bring many limitations in adoption while separating explanations from the model can afford several beneficial flexibilities \cite{ribeiro2016modelagnostic}. As a result, model-agnostic methods have been devised to provide interpretations without knowing model details.




\textbf{Partial Dependence Plot: } Partial Dependence Plots (PDP) reveal the dependence between the target function and several target features. The partial function $ \hat{f_{xs}}(x_s) $ is estimated by calculating averages in the training data, also known as the Monte Carlo method. After setting up a grid for the features we are interested in (target features), we set all target features in our training set to be the value of grid points, then make predictions and average them all at each grid. The drawback of PDP is that one target feature produces 2D plots and two produce 3D plots while it can be pretty hard for a human to understand plots in higher dimensions.

\begin{equation}
    \label{eq:1}
    \hat{f}_{xs}(x_s) = \frac{1}{n}\sum_{1}^{n}\hat{f}(x_s, x_c^{i})
\end{equation}

\textbf{Individual Conditional Expectation: } Individual Conditional Expectation (ICE) is similar to PDP. The difference is that PDP calculates the average over the marginal distribution while ICE keeps them all. Each line in the ICE plot represents predictions for each individual. Without averaging on all instances, ICE unveils heterogeneous relationships but is limited to only one target feature since two features result in overlay surfaces that cannot be identified by human eyes \cite{molnar2019}.

\textbf{Accumulated Local Effects: } Accumulated Local Effects (ALE) averages the changes in the predictions and accumulate them over the local grid. The difference with PDP is that the value at each point of the ALE curve is the difference to the mean prediction calculated in a small window rather than all of the grid. Thus ALE eliminates the effect of correlated features \cite{molnar2019} which makes it more suitable in healthcare because it's usually irrational to assume young people having similar physical conditions with the elderly.


\textbf{Permutation Feature Importance: } The idea behind Permutation Feature Importance is intuitive. A feature is significant for the model if there is a noticeable increase in the model's prediction error after permutation. On the other hand, the feature is less important if the prediction error remains nearly unchanged after shuffling.

\textbf{Local Interpretable Model-agnostic Explanations:} Local Interpretable Model-agnostic Explanations (LIME) uses interpretable models to approximate the predictions of the original black-box model in specific regions. LIME works for tabular data, text, and images, but the explanations may not be stable enough for medical applications.

\textbf{Shapley Additive Explanation: } Shapley Additive exPlanation (SHAP) borrows the idea of Shapley value from Game Theory \cite{shapley_1953}, which represents contributions of each player in a game. Calculating Shapley values is computationally expensive when there are hundreds of features, thus Lundberg, Scott M., and Su-In Lee proposed a fast implementation for tree-based models to boost the calculation process \cite{lundberg2017unified}. SHAP has a solid theoretical foundation but is still computationally slow for a lot of instances.

To summarize, PDP, ICE, and ALE only use graphs to visualize the impact of different features while Permutation Feature Importance, LIME, and SHAP provide numerical feature importance that quantitatively ranks the importance of each feature.


\subsection{\textbf{Metrics for Interpretability Evaluation}}

Different interpretation methods try to find out the most important features to provide explanations for the output. But as Doshi-Velez and Kim questioned, "Are all models in all defined-to-be-interpretable model classes equally interpretable?"  \cite{doshivelez2017rigorous} And how can we measure the quality of different interpretation methods? 


\textbf{Faithfulness:} Faithfulness incrementally removes each of the attributes deemed important by the interpretability metric, and evaluate the effect on the performance. Then it calculates the correlation between the weights (importance) of the attributes and corresponding model performance and returns correlation between attribute importance weights and the corresponding effect on classifier \cite{alvarezmelis2018robust}.

\textbf{Monotonicity:} Monotonicity incrementally adds each attribute in order of increasing importance. As each feature is added, the performance of the model should correspondingly increase, thereby resulting in monotonically increasing model performance, and it returns True of False \cite{luss2019generating}.
\color{black}

In our experiment, both faithfulness and monotonicity are employed to evaluate the interpretation of different machine learning models. 

\section{Empirical Study on COVID}

In this section, features in our raw dataset and procedures of data preprocessing are introduced. After preprocessing, four different models: decision tree, random forest, gradient boosted trees, and neural networks are trained on the dataset. Model interpretation is then employed to understand how different models make predictions, and patients that models make false diagnoses are investigated respectively.

\subsection{\textbf{Dataset and Perprocessing}}

The raw dataset consists of patients with confirmed SARS-CoV-2 laboratory tests between 18th Jan. 2020 and 5th Mar. 2020, in Zhuhai, China. Our Research Ethics Committee waived written informed consent for this retrospective study that evaluated de-identified data and involved no potential risk to patients. All the data of patients have been anonymized before analysis.

Tables in the Appendix list all 74 features in the raw dataset consisting of Body Mass Index (BMI), Complete Blood Count (CBC), Blood Biochemical Examination, inflammatory markers, symptoms, anamneses, among others. Whether or not health care professionals will order a test for patients is based on various factors such as medical history, physical examination, and etc. Thus, there is no standard set of tests that are compulsory for every individual which introduces data sparsity. For instance, Left Ventricular Ejection Fraction (LVEF) are mostly empty because most patients are not required to take the color doppler ultrasound test .

\color{black}
After pruning out irrelevant features, such as patients' medical numbers that provide no medical information, and features that have no patients' records (no patient took this test), 86 patients' records with 55 features are selected for further investigation. Among those, 77 records are used for training, cross-validation, and 9 reserved for testing. The feature for classification is Severity01 which indicates normal with 0, and severe with 1. More detailed descriptions about features in our dataset are listed in the Appendix.

Feature engineering is applied before training and interpreting our models, as some features may not provide valuable information or provide redundant information.

First, constant and quasi-constant features were removed. For instance, the two features, PCT2 and Stomachache, have the same value for all patients providing no valuable information in distinguishing normal and severe patients.


Second, correlated features were removed because they provide redundant information. Table \ref{tab:feature_correlation} lists all correlated features using Pearson’s correlation coefficient. 


\begin{table}[H]
\centering
\caption{Feature Correlation}
\begin{tabular}{@{}ccc@{}}
\toprule
Feature 1    & Feature 2 & Correlation\\ \midrule
cTnICKMBOrdinal1 & cTnICKMBOrdinal2             & 0.853741       \\
LDH & HBDH             &  0.911419   \\
NEU2           & WBC2             &  0.911419  \\
LYM2       & LYM1             & 0.842688    \\ 
NTproBNP       & N2L2             & 0.808767   \\ 
BMI       & Weight             & 0.842409    \\  
NEU1       & WBC1             & 0.90352    \\  \bottomrule
\label{tab:feature_correlation}
\end{tabular}
\end{table}

\begin{enumerate}  
\item There is strong correlation between cTnICKMBOrdinal1 and cTnICKMBOrdinal2 because they are the same test among a short range of time which is the same for LYM1 and LYM2.
\item LDH and HBDH levels are significantly correlated with heart diseases, and the HBDH/LDH ratio can be calculated to differentiate between liver and heart diseases.
\item Neutrophils (NEU1/NEU2) are all correlated to the immune system. In fact, most of the white blood cells that lead the immune system’s response are neutrophils. Thus, there is a strong correlation between NEU1 and WBC1, NEU2 and WBC2.
\item In the original dataset, there is no much information about N2L2 which is correlated with NTproBNP, thus NTproBNP remains.
\item the correlation between BMI and weight is straight forward because Body Mass Index (BMI) is a person’s weight in kilograms divided by the square of height in meters.
\end{enumerate}

Third, statistical methods that calculate mutual information is employed to remove features with redundant information.

\begin{table}[H]
\centering
\caption{Features with Mutual Information}
\begin{tabular}{@{}cc@{}}
\toprule
Statistical Methods & Removed Features \\ \midrule
Mutual Information & Height, CK, HiCKMB, Cr, WBC1, Hemoptysis      \\
Univariate & Weight, AST, CKMB, PCT1, WBC2        \\  \bottomrule
\end{tabular}
\end{table}

Mutual information is calculated using equation \ref{eq:2} that determines how similar the joint distribution p(X, Y) is to the products of individual distributions p(X)p(Y). Univariate Test measures the dependence of two variables, and a high p-value indicates a less similar distribution between X and Y. 

\begin{equation}
    I(X;Y)=\sum_{x,y}p(x, y)log\frac{p(x,y)}{p(x)p(y)} \label{eq:2}
\end{equation}

\par
After feature engineering, there are 37 features left for training and testing.

\subsection{\textbf{Training Models}}

Machine learning models outperform humans in many different areas in terms of accuracy. Interpretable models such as the decision tree are easy to understand, but not suitable for large scale applications. Complex models achieve high accuracy while giving less explanation. 

For healthcare applications, both accuracy and interpretability are significant. Four different models are selected to extract information from our dataset: Decision Tree, Random Forests, Gradient Boosted Trees, and Neural Networks.

\begin{figure*}
\centering
\begin{subfigure}[b]{0.475\textwidth}
    \centering
    \includegraphics[width=\textwidth]{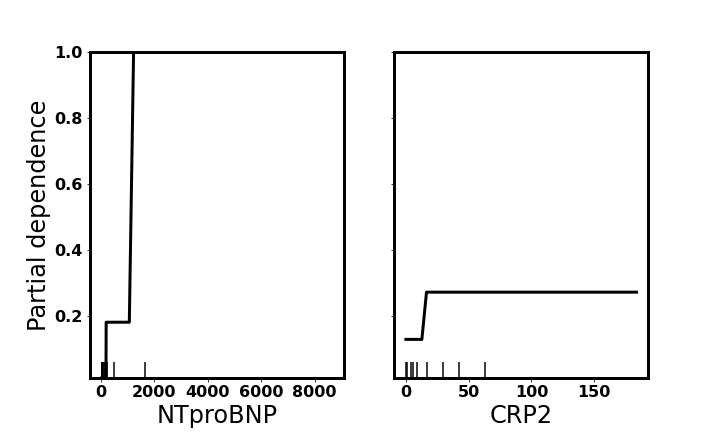}
    \caption{Decision Tree}
    \label{fig:dt_pdp}
\end{subfigure}
\begin{subfigure}[b]{0.475\textwidth}
    \centering
    \includegraphics[width=\textwidth]{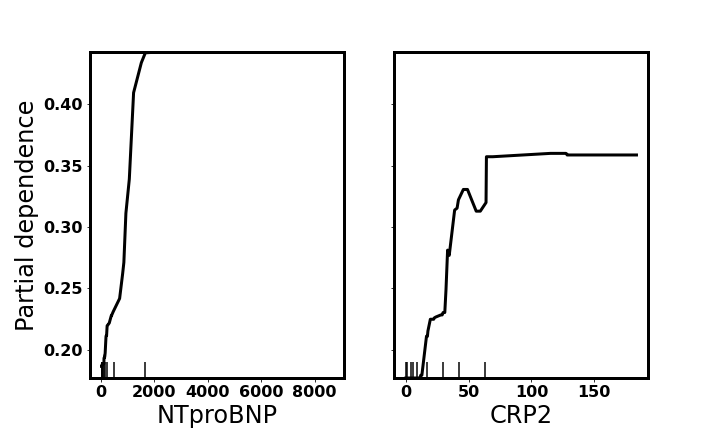}
    \caption{Random Forest}
    \label{fig:rf_pdp}
\end{subfigure}
\begin{subfigure}[b]{0.475\textwidth}
    \centering
    \includegraphics[width=\textwidth]{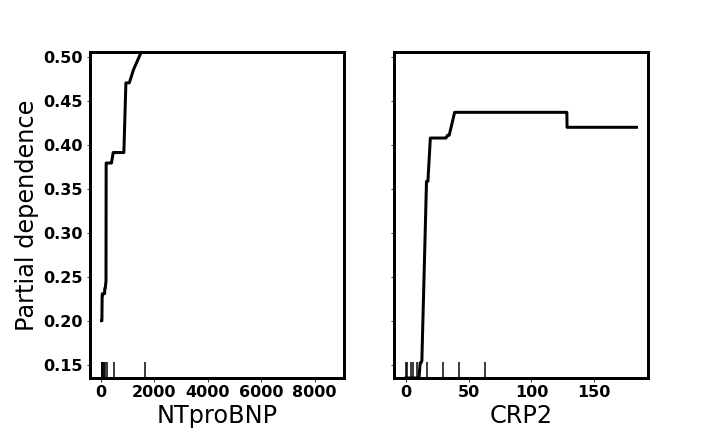}
    \caption{Gradient Boosted Trees}
    \label{fig:gbt_pdp}
\end{subfigure}
\begin{subfigure}[b]{0.475\textwidth}
    \centering
    \includegraphics[width=\textwidth]{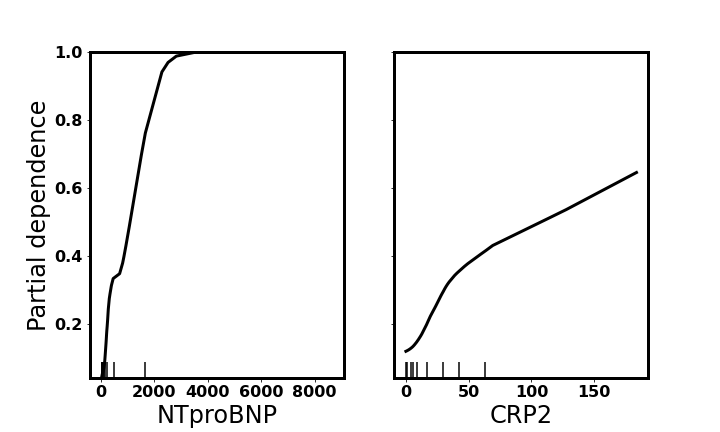}
    \caption{Neural Networks}
    \label{fig:nn_pdp}
\end{subfigure}
\caption{Partial Dependence Plot: There is a positive correlation between the level of NTproBNP/CRP and the probability of turning severe because as NTproBNP/CRP increases, the average possibility (y-axis) of turning severe increases.}
\label{fig.pdp}
\end{figure*}

\textbf{Decision Tree:} Decision Tree (DT) is a widely adopted method for both classification and regression. It's a non-parametric supervised learning method that infers decision rules from data features. The decision tree try to find decision rules that make the best split measured by Gini impurity or entropy. More importantly, the generated decision tree can be visualized, thus easy to understand and interpret \cite{Breiman2015Classification}.

\textbf{Random Forest:} Random Forests (RF) is a kind of ensemble learning method \cite{Breiman2001Random} that employs bagging strategy. Multiple decision trees are trained using the same learning algorithm, and then predictions are aggregated from the individual decision tree. Random forests produce great results most of the time even without much hyper-parameter tuning. As a result, it has been widely accepted for its simplicity and good performance. However, it is rather difficult for humans to interpret hundreds of decision trees, so the model itself is less interpretable than a single decision tree.

\textbf{Gradient Boosted Trees:} Gradient Boosted Trees is another ensemble learning method that employs boosting strategy \cite{gbtrees}. Through sequentially adding one decision tree at one time, gradient boosted trees combine results along the way. With fine-tuned parameters, gradient boosting can result in better performance than random forests. Still, it is tough for humans to interpret a sequence of decision trees and thus considered as black-box models.

\textbf{Neural Networks:} Neural Networks could be the most promising model in achieving a high accuracy and even outperforms humans in medical imaging \cite{maier2018gentle}. Though the whole network is difficult to understand, deep neural networks are stacks of simple layers, thus can be partially understood through visualizing outputs of intermediate layers \cite{Montavon_2018}.


As for the implementation, there is no hyperparameter for the decision tree. For random forests, 100 trees are used during the initialization. The hyperparameters for gradient boosted trees are selected according to prior experience. The structure for neural networks is listed in table \ref{tab:nn}. All these methods are implemented using scikit-learn \cite{scikit-learn}, Keras and python3.6.

\begin{table}[H]
\centering
\caption{The structure of Neural Networks}
\begin{tabular}{@{}ccc@{}}
\toprule
Layer Type   & Output Shape & Param \\ \midrule
Dense & (None, 10) & 370           \\
Dropout & (None, 10) & 0           \\
Dense & (None, 15) & 165           \\
Dropout & (None, 15) & 0           \\
Dense & (None, 5) & 80           \\
Dropout & (None, 5) & 0           \\
Dense & (None, 1) & 6           \\
\bottomrule
\label{tab:nn}
\end{tabular}
\end{table}

After training, gradient boosted trees and neural networks achieve the highest precision on the test set. Among 9 patients in our test set, four of them are severe. Both the decision tree and random forests fail to identify two severe patients, while Gradient Boosted Trees and Neural Networks find all of the severe patients. 

\begin{table}[H]
\centering
\caption{Classification Results on our dataset}
\begin{tabular}{p{2cm}p{1.1cm}cccp{1cm}}
\toprule
Classifier    & CV & \multicolumn{3}{c}{Test Set} & 95\% confidence interval \\ \midrule
                        & F1     & Precision  & Recall  & F1    &\\
Decision Tree           & 0.55   & 0.67       & 0.50    & 0.57  & 0.31\\
Random Forest           & 0.62   & 0.67       & 0.50    & 0.57  & 0.31\\
Gradient Boosted Trees  & 0.67   & 0.78       & 1.00    & 0.80  & 0.27\\
Neural Networks         & 0.58   & 0.78       & 1.00    & 0.80  & 0.27\\ \bottomrule
\end{tabular}
\end{table}


\begin{figure*}
\centering
\begin{subfigure}[b]{0.485\textwidth}
    \centering
    \includegraphics[width=\textwidth]{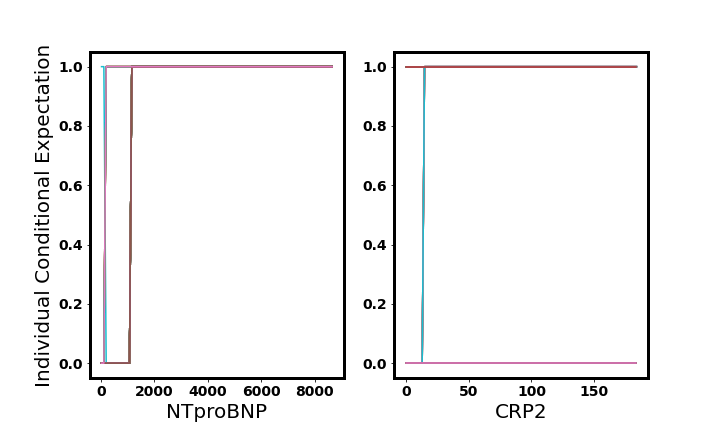}
    \caption{Decision Tree}
    \label{fig:dt_ice}
\end{subfigure}
\hfill
\begin{subfigure}[b]{0.485\textwidth}
    \centering
    \includegraphics[width=\textwidth]{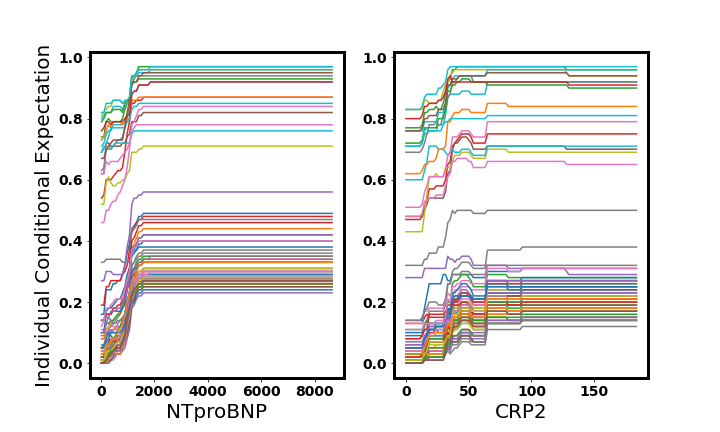}
    \caption{Random Forest}
    \label{fig:rf_ice}
\end{subfigure}
\hfill
\begin{subfigure}[b]{0.485\textwidth}
    \centering
    \includegraphics[width=\textwidth]{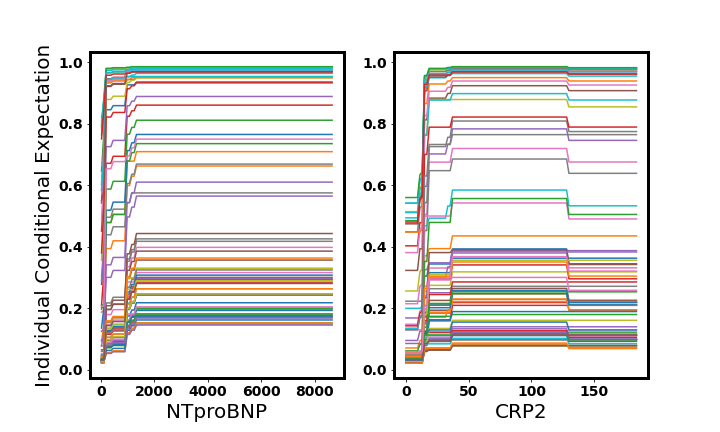}
    \caption{Gradient Boosted Trees}
    \label{fig:gbt_ice}
\end{subfigure}
\hfill
\begin{subfigure}[b]{0.485\textwidth}
    \centering
    \includegraphics[width=\textwidth]{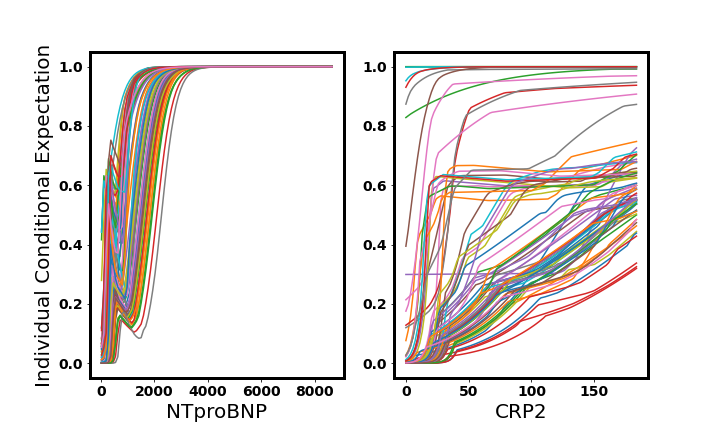}
    \caption{Neural Networks}
    \label{fig:nn_ice}
\end{subfigure}
\caption{Individual Conditional Expectation: Each line in different colors represents a patient. As we increase NTproBNP/CRP while keeping other features the same, the probability of turning severe increases for each individual, but each patient has a different starting level because their other physical conditions differ.}
\label{fig.ice}
\end{figure*}


\begin{figure*}
\centering
\begin{subfigure}[b]{0.485\textwidth}
    \centering
    \includegraphics[width=\textwidth]{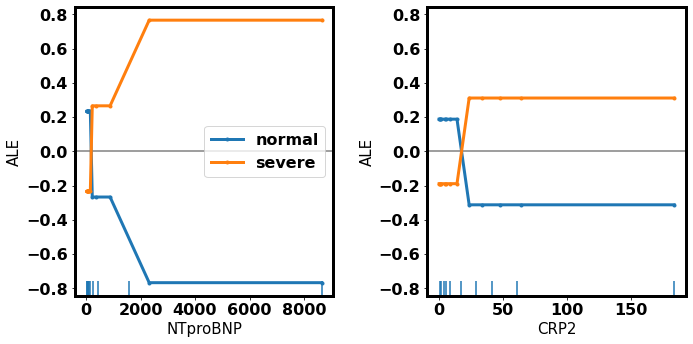}
    \caption{Decision Tree}
    \label{fig:dt_ale}
\end{subfigure}
\hfill
\begin{subfigure}[b]{0.485\textwidth}
    \centering
    \includegraphics[width=\textwidth]{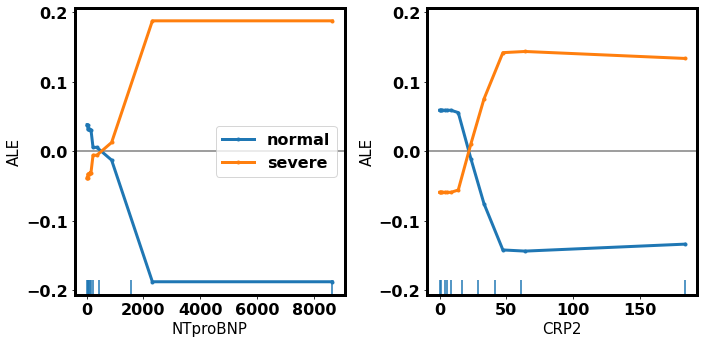}
    \caption{Random Forest}
    \label{fig:rf_ale}
\end{subfigure}
\hfill
\begin{subfigure}[b]{0.485\textwidth}
    \centering
    \includegraphics[width=\textwidth]{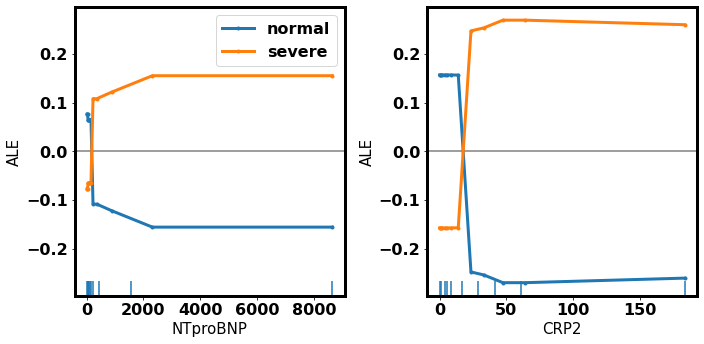}
    \caption{Gradient Boosted Trees}
    \label{fig:gbt_ale}
\end{subfigure}
\hfill
\begin{subfigure}[b]{0.485\textwidth}
    \centering
    \includegraphics[width=\textwidth]{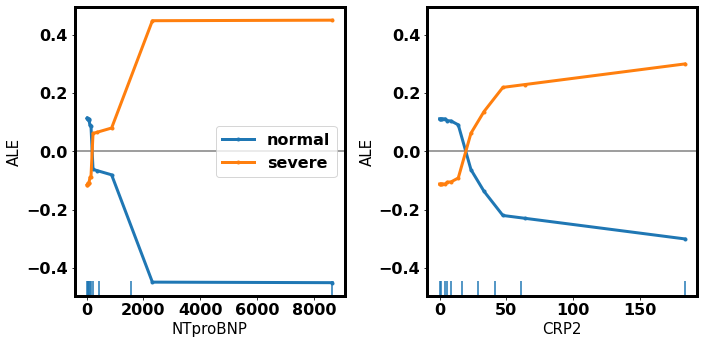}
    \caption{Neural Networks}
    \label{fig:nn_ale}
\end{subfigure}
\caption{Accumulated Local Effects: As the level of NTproBNP/CRP increases, the possibility of turning severe (yellow) goes above the average.}
\label{fig.ale}
\end{figure*}

\newpage
\subsection{\textbf{Interpretation (Permutation Feature Importance)}}

First, we use  Permutation Feature Importance to find the most important features in different models. In table \ref{tab:important_feature}, CRP2 and NTproBNP are recognized as most important features by most models.

\begin{table}[H]
\centering
\caption{Five most important features}
\begin{tabular}{@{}ccc@{}}
\toprule
Model                   & Most Important Features \\ \midrule
Decision Tree           & NTproBNP, CRP2, ALB2, Temp, Symptom      \\
Random Forest           & CRP2, NTproBNP, cTnI, LYM1, ALB2         \\
Gradient Boosted Trees  & CRP2, cTnITimes, LYM1, NTproBNP, Phlegm  \\
Neural Networks         & NTproBNP, CRP2, CRP1, LDH, Age           \\  \bottomrule
\label{tab:important_feature}
\end{tabular}
\end{table}

According to medical knowledge, CRP refers to C-Reactive Protein, which increases when there's inflammation or viral infection in the body. C-reactive protein (CRP) levels are positively correlated with lung lesions and could reflect disease severity\cite{WANG2020332}. NTproBNP refers to N-Terminal prohormone of Brain Natriuretic Peptide, which will be released in response to changes in pressure inside the heart. The CRP level in severe patients rises due to viral infection, and patients with higher NT-proBNP (above 88.64 pg/mL) level had more risks of in-hospital death \cite{Gao2020.03.07.20031575}.


\newpage
\subsection{\textbf{Interpretation (PDP, ICE, ALE)}}

After recognizing the most important features, PDP, ICE, and ALE are employed to further visualize the relationship between CRP and NTproBNP.

In the PDPs, all of the four models indicate a higher risk of turning severe with the increase of NTproBNP and CRP which is consistent with the retrospective study on COVID-19. The difference is that different models have different tolerances and dependence on NTproBNP and CRP. Averagely, the decision tree has less tolerance on a high level of NTproBNP ($>$2000ng/ml), and gradient boosted trees give a much higher probability of death as CRP increases. Since PDPs only calculate an average of all instances, we use ICEs to identify heterogeneity.
    
ICE reveals individual differences. Though all of the models give a prediction of a higher risk of severe as NTproBNP and CRP increase, some patients have a much higher initial probability which indicates other features have an impact on overall predictions. For example, elderly people have higher NTproBNP than young people and have a higher risk of turning severe.

In the ALEs,  as NTproBNP and CRP get higher, all of the four models give a more positive prediction of turning severe, which coincides with medical knowledge.

\clearpage
\subsection{\textbf{Misclassified Patients}}

Even though the most important features revealed by our models exhibit medical meaning, some severe patients fail to be recognized. Both Gradient Boosted Trees and Neural Networks recognize all severe patients and yield a recall of 1.00, while the decision tree and random forests fail to reveal two of them. 

Patient No. 2 (normal) is predicted with a probability of 0.53 of turning severe which is around the boundary (0.5). While for patient No. 5 (severe), the model gives a relatively low probability of turning severe (0.24).

\begin{table}[H]
\centering
\caption{Misclassified Patients}
\begin{tabular}{ccccc}
\toprule
No & Class  & Probability of Severe & Prediction & Type\\ \midrule
2 & Normal & 0.53 & Severe & \textbf{False Positive}       \\
5 & Severe & 0.24  & Normal & \textbf{False Negative}   \\ \bottomrule
\end{tabular}
\label{tab:misclassified}
\end{table}

\subsection{\textbf{Interpretation (False Negative)}}

Suppose different models represent different doctors, then the decision tree and random forests make the wrong diagnosis for patient no. 5. The reason human doctors classified the patient as severe is that he actually needed a respirator to survive. To further investigate why the decision tree and random forests make wrong predictions, Local Interpretable Model-agnostic Explanations (LIME) and (Shapley Additive Explanation) SHAP are employed.

LIME: Features in green have a positive contribution to the prediction (increasing the probability of turning severe), and features in red have a negative effect on the prediction (decreasing the probability of turning severe).

SHAP: Features pushing the prediction to be higher (severe) are shown in red,  and those pushing the prediction to be lower (normal) are in blue.

\subsubsection{\textbf{Wrong Diagnoses}}

\begin{figure*}
\centering
\begin{subfigure}[b]{0.45\textwidth}
    \centering
    \includegraphics[width=\textwidth]{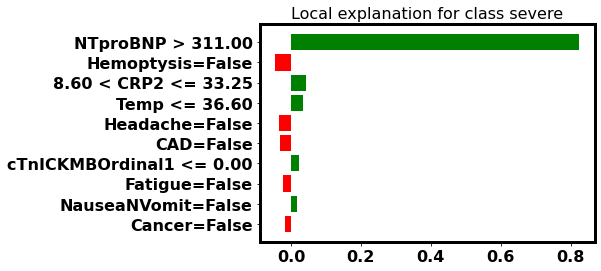}
    \caption{Decision Tree}
    \label{fig:lime_dt_5}
\end{subfigure}
\hfill
\begin{subfigure}[b]{0.45\textwidth}
    \centering
    \includegraphics[width=\textwidth]{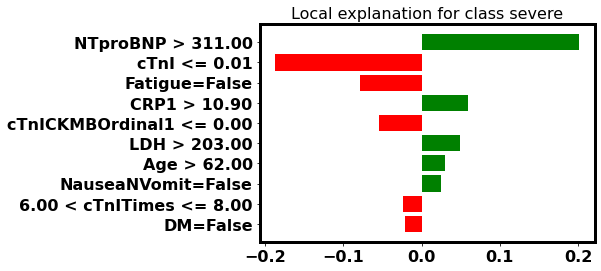}
    \caption{Random Forests}
    \label{fig:lime_rf_5}
\end{subfigure}
\hfill
\begin{subfigure}[b]{0.45\textwidth}
    \centering
    \includegraphics[width=\textwidth]{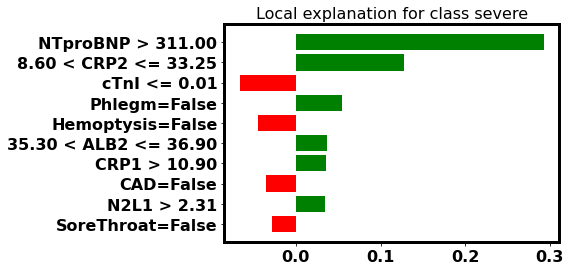}
    \caption{Gradient Boosted Trees}
    \label{fig:lime_xgbc_5}
\end{subfigure}
\hfill
\begin{subfigure}[b]{0.45\textwidth}
    \centering
    \includegraphics[width=\textwidth]{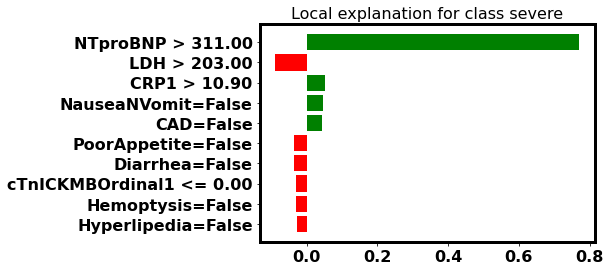}
    \caption{Neural Networks}
    \label{fig:lime_nn_5}
\end{subfigure}
\caption{LIME Explanation (\textbf{False-Negative Patient No.5}): Features in green have a positive contribution to the prediction (increasing the probability of turning severe), and features in red have a negative effect on the prediction (decreasing the probability of turning severe)}
\label{fig:lime_5}
\end{figure*}

Take the decision tree as an example, in the figure \ref{fig:lime_dt_5}, the explanation by LIME illustrates that NTproBNP and CRP are two features (in green) that have a positive impact on the probability of turning severe. Even though patient No.5 is indeed severe, the decision tree gives an overall prediction of normal (false negative). Thus, we would like to investigate features that have a negative impact on the probability of turning severe.

In the figure \ref{fig:shap_xgbc_5}, the explanation by SHAP reveals that the patient is diagnosed as normal by the decision tree because the patient has no symptom. Even though the patient has a high NTproBNP and CRP, having no symptom makes it less likely to classify him as severe. The record was taken when the patient came to the hospital for the first time. It is likely that the patient developed symptoms later and turned severe.

However, both gradient boosted trees and neural networks are not deceived by the fact the patient has no symptom. Their predictions indicate that the patient is likely to turn severe in the future.

\subsubsection{\textbf{Correct Diagnoses}}

\begin{figure*}
\centering
\begin{subfigure}[b]{1.0\textwidth}
    \centering
    \includegraphics[width=\textwidth]{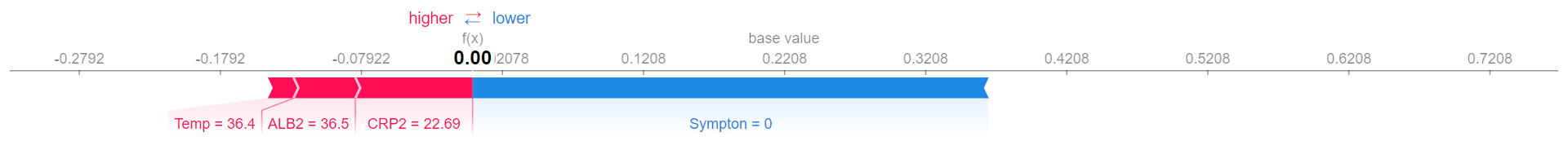}
    \caption{Decision Tree}
    \label{fig:shap_dt_5}
\end{subfigure}
\hfill
\begin{subfigure}[b]{1.0\textwidth}
    \centering
    \includegraphics[width=\textwidth]{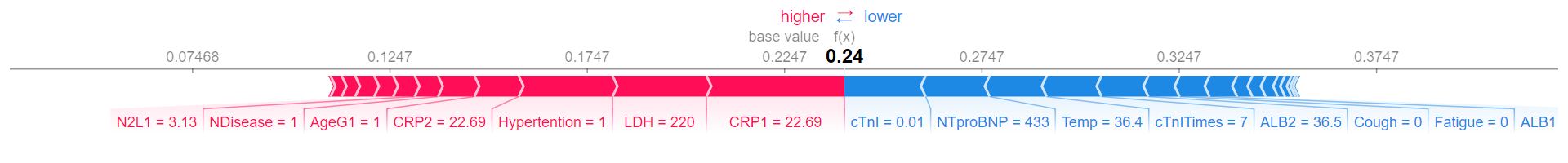}
    \caption{Random Forests}
    \label{fig:shap_rf_5}
\end{subfigure}
\hfill
\begin{subfigure}[b]{1.0\textwidth}
    \centering
    \includegraphics[width=\textwidth]{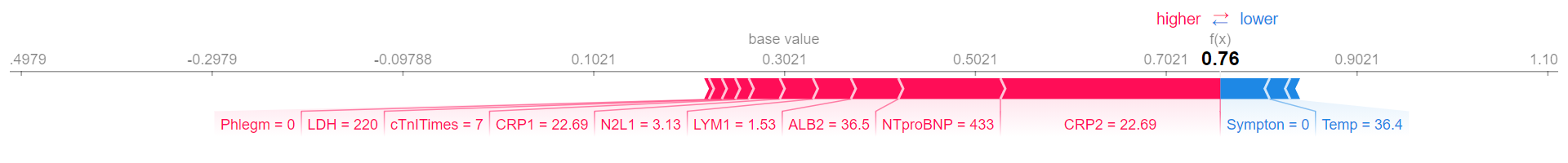}
    \caption{Gradient Boosted Trees}
    \label{fig:shap_xgbc_5}
\end{subfigure}
\hfill
\begin{subfigure}[b]{1.0\textwidth}
    \centering
    \includegraphics[width=\textwidth]{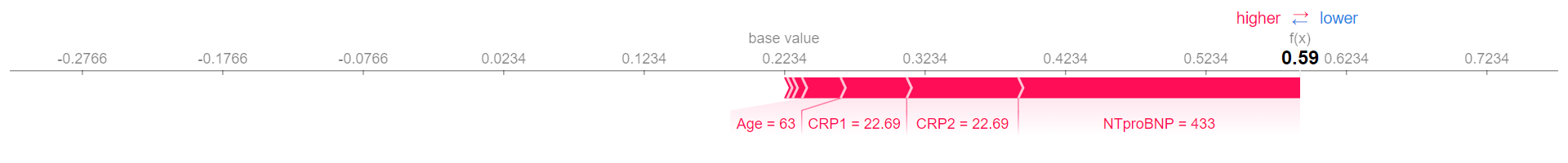}
    \caption{Neural Networks}
    \label{fig:shap_nn_5}
\end{subfigure}
\hfill
\caption{SHAP Explanation (\textbf{False-Negative Patient No.5}): Features pushing the prediction to be higher (severe) are shown in red,  and those pushing the prediction to be lower (normal) are in blue.}
\label{fig:shap_5}
\end{figure*}

In the figure \ref{fig:shap_xgbc_5} and figure \ref{fig:shap_nn_5}, gradient boosted trees and neural networks do not prioritise the feature symptom. They put more weight on test results (NTproBNP and CRP). Thus they make correct predictions based on the fact that the patient's test results are serious.

Besides, neural networks notice that the patient is elderly (Age = 63). If we calculate the average age in different severity levels, it is noticeable that elderly people are more likely to deteriorate.

\begin{table}[H]
\centering
\caption{Average Age in different severity levels}
\begin{tabular}{@{}cc@{}}
\toprule
Severity Level   & Average Age\\ 
\midrule
0      &          36.83 \\ 
1      &          47.45 \\ 
2      &          54.31 \\ 
3      &          69.40 \\ 
\bottomrule
\end{tabular}
\end{table}

Gradient boosted trees and neural networks make correct predictions because they trust more in test results, while the decision tree relies more on whether or not a patient has symptoms. As a result, gradient boosted trees and neural networks are capable of recognizing patients that are likely to turn severe in the future while the decision tree makes predictions relying more on patients' current situation.

Medical research is a case-by-case study. Every patient is unique. It's strenuous to find a single criterion that suits every patient, thus it's important to focus on each patient and make a diagnosis accordingly. This is one of the benefits of using interpretable machine learning. It unveils the most significant features for most patients and provides the interpretation for each patient as well.
\color{black}

\subsection{\textbf{Interpretation (False Positive)}}

With limited medical resources at the initial outbreak of the pandemic, it's equally important to investigate false positive, so that valuable resources can be distributed to patients in need.

In table \ref{tab:misclassified}, patient 2 is normal, but all of our models diagnose the patient as severe. To further explain the false positive prediction, table \ref{tab:normal2} lists anonymized medical records for patient 2 (normal) and patient 5 (severe) for comparison.

\begin{table}[H]
\centering
\caption{Record of the false positive Patient 2}
\begin{tabular}{cp{2.2cm}p{2.1cm}}
\toprule
Feature   & Patient 5 (Severe) & Patient 2 (Normal)\\ 
\midrule
Sex                 &   1.00    &   1.00 \\
Age                 &   63.00   &   42.00 \\
AgeG1               &   1.00    &   0.00 \\
\textbf{Temp}       &   36.40   &   \textbf{37.50} \\
cTnITimes           &   7.00    &   8.00 \\
cTnI                &   0.01    &   0.01 \\
cTnICKMBOrdinal1    &  0.00     &   0.00 \\
\textbf{LDH}        &  220.00   &   \textbf{263.00} \\
\textbf{NTproBNP}   &  433.00   &   \textbf{475.00} \\
\textbf{LYM1}       &  1.53     &   \textbf{1.08} \\
N2L1                &  3.13     &   2.16 \\
\textbf{CRP1}       &  22.69    &   \textbf{36.49} \\
ALB1                &  39.20    &   37.60 \\
\textbf{CRP2}       &  22.69    &   \textbf{78.76} \\
ALB2                &  36.50    &   37.60 \\
Symptoms            &  None     &   Fever \\
NDisease            & Hypertention & Hypertention, DM, Hyperlipedia \\
\bottomrule
\label{tab:normal2}
\end{tabular}
\end{table}

\newpage
\subsubsection{\textbf{Doctors' Diagnoses}}

We present the test results of both patients to doctors without indicating which patient is severe. All doctors mark patient No. 2 as more severe which is the same as our models. Doctors' decisions are based on the COVID-19 Diagnosis and Treatment Guide in China. The increased level in CRP, LDH, decreased level in LYM are associated with severe COVID-19 infection in the guideline, and patient 2 has a higher level of CRP and LDH, a lower level of LYM than patient 5. As a result, doctors' diagnoses are consistent with models' predictions

\subsubsection{\textbf{Models' Diagnoses}}

\begin{figure*}
\centering
\begin{subfigure}[b]{0.45\textwidth}
    \centering
    \includegraphics[width=\textwidth]{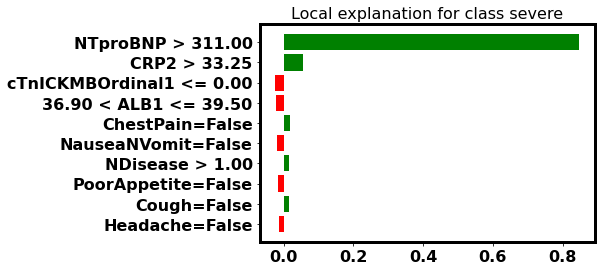}
    \caption{Decision Tree}
    \label{fig:lime_dt_2}
\end{subfigure}
\hfill
\begin{subfigure}[b]{0.45\textwidth}
    \centering
    \includegraphics[width=\textwidth]{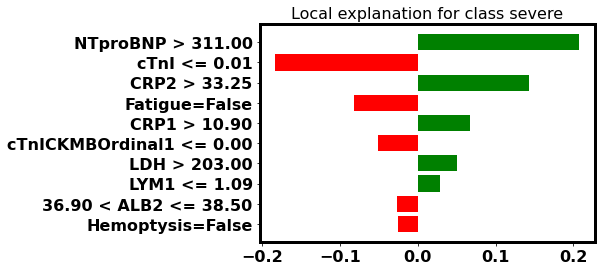}
    \caption{Random Forests}
    \label{fig:lime_rf_2}
\end{subfigure}
\hfill
\begin{subfigure}[b]{0.45\textwidth}
    \centering
    \includegraphics[width=\textwidth]{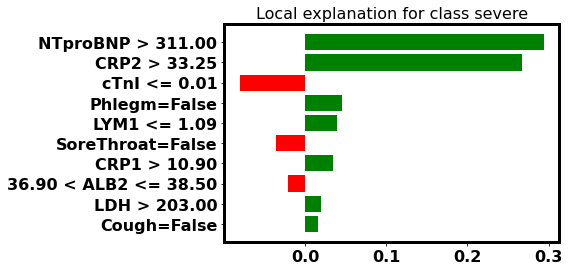}
    \caption{Gradient Boosted Trees}
    \label{fig:lime_xgb_2}
\end{subfigure}
\hfill
\begin{subfigure}[b]{0.45\textwidth}
    \centering
    \includegraphics[width=\textwidth]{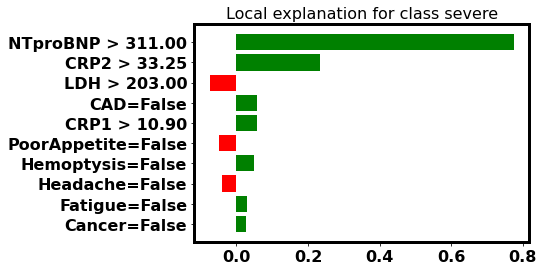}
    \caption{Neural Networks}
    \label{fig:lime_nn_2}
\end{subfigure}
\hfill
\caption{LIME Explanation (\textbf{False-Positive Patient No.2}): Features in green have a positive contribution to the prediction (increasing the probability of turning severe), and features in red have a negative effect on the prediction (decreasing the probability of turning severe)}
\label{fig:lime_2}
\end{figure*}

\begin{figure*}
\centering
\begin{subfigure}[b]{1.0\textwidth}
    \centering
    \includegraphics[width=\textwidth]{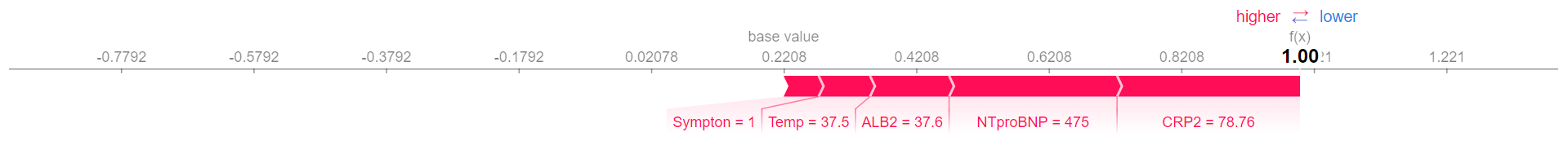}
    \caption{Decision Tree}
    \label{fig:shap_dt_2}
\end{subfigure}
\hfill
\begin{subfigure}[b]{1.0\textwidth}
    \centering
    \includegraphics[width=\textwidth]{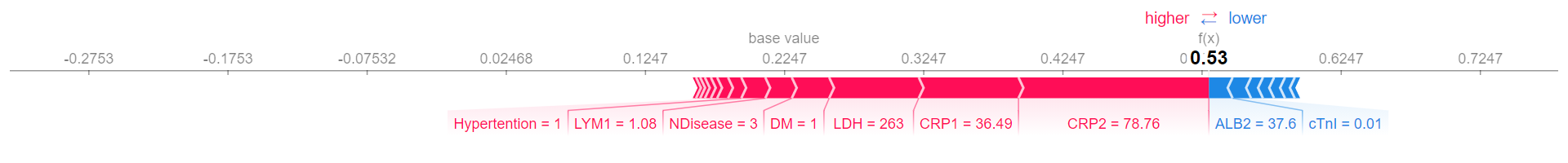}
    \caption{Random Forests}
    \label{fig:shap_rf_2}
\end{subfigure}
\hfill
\begin{subfigure}[b]{1.0\textwidth}
    \centering
    \includegraphics[width=\textwidth]{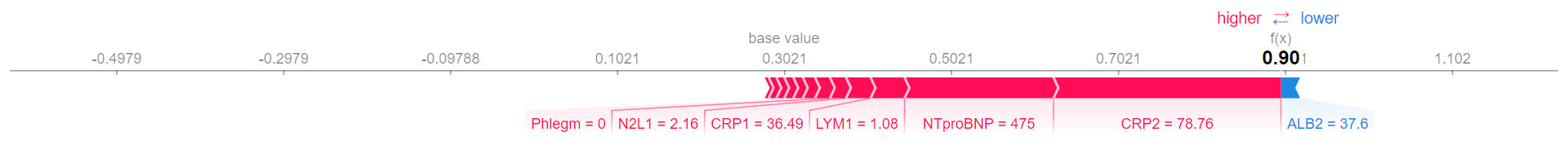}
    \caption{Gradient Boosted Trees}
    \label{fig:shap_xgbc_2}
\end{subfigure}
\hfill
\begin{subfigure}[b]{1.0\textwidth}
    \centering
    \includegraphics[width=\textwidth]{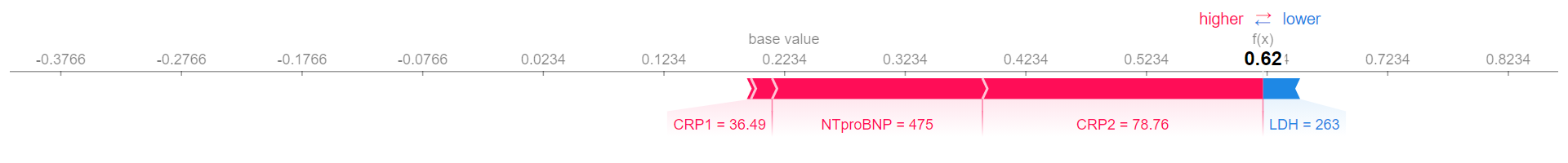}
    \caption{Neural Networks}
    \label{fig:shap_nn_2}
\end{subfigure}
\hfill
\caption{SHAP Explanation (\textbf{False-Positive Patient No.2}): Features pushing the prediction to be higher (severe) are shown in red,  and those pushing the prediction to be lower (normal) are in blue.}
\label{fig:shap_2}
\end{figure*}

Even though all of the four models make the same predictions as human doctors, it's important to confirm models' predictions are in accordance with medical knowledge. Table \ref{tab:important_feature_patient} lists the three most important features in the interpretation of LIME and SHAP. More detailed interpretations are illustrated in the figure \ref{fig:lime_2} and figure \ref{fig:shap_2}.

\begin{table}[H]
\centering
\caption{Most important features from LIME, SHAP}
\begin{tabular}{@{}cc@{}}
\toprule
Model                   & LIME                          \\  
\toprule
Decision Tree           & NTproBNP, CRP2, NauseaNVomit  \\
Random Forest           & NTproBNP, CRP2, CRP1          \\  
Gradient Boosted Trees  & NTproBNP, CRP2, LYM1          \\  
Neural Networks         & NTproBNP, CRP2, PoorAppetite  \\
\toprule
Model                   & SHAP  \\ 
\toprule
Decision Tree           & CRP2, NTproBNP, ALB2 \\
Random Forests          & CRP2, CRP1, LDH      \\
Gradient Boosted Trees  & CRP2, NTproBNP, LDH  \\
Neural Networks         & CRP2, NTproBNP, CRP1 \\  
\bottomrule
\label{tab:important_feature_patient}
\end{tabular}
\end{table}






In table \ref{tab:important_feature_patient}, NTproBNP, CRP, LYM, LDH are the most common features that are deemed crucial by all different models. The three features, CRP, LYM, LDH, are listed as the most indicative biomarkers in the COVID-19 guideline. While the correlation between NTproBNP and COVID-19 are investigated in a paper from World Health Organization (WHO) global literature on coronavirus disease, that reveals elevated NTproBNP is associated with increased mortality in patients with COVID-19 \cite{Pranata387}.

As a result, the prediction of false-positive is consistent with doctors' diagnoses. Patient 2 who is normal is diagnosed as severe by both doctors and models. One possibility is that even though the patients' test results are not optimistic, he did not require a respirator to survive when he came to the hospital for the first time, so he was classified as normal. In this way, models' predictions can act as a warning. If a patient is diagnosed as severe by models, and the prediction is in accordance with medical knowledge, but the patient feels normal, we can suggest to the patient to put more attention on his health condition.

In conclusion, as illustrated previously in the explanation for patient 5 (false negative), every patient is unique. Some patients are more resistant to viral infection, while some are more vulnerable. Pursuing a perfect model is tough in healthcare, but we can try to understand how different models make predictions using interpretable machine learning to be more responsible with our diagnoses.
\color{black}

\newpage
\subsection{\textbf{Evaluating Interpretation}}

Though we do find some indicative symptoms of COVID-19 through model interpretation, they are confirmed credible because these interpretations are corroborated by medical research. If we use the interpretation to understand a new virus at the early stage of an outbreak, there will be less evidence to support our interpretation. Thus we use Monoitinicity and Faithfulness to evaluate different interpretations using IBM AIX 360 toolbox \cite{aix360-sept-2019}. The decision tree only provides a binary prediction (0 or 1) rather than a probability between 0 and 1, so it cannot be evaluated using Monotonicity and Faithfulness.

\begin{table}[H]
\centering
\caption{Failthfulness Evaluation}
\begin{tabular}{@{}ccc@{}}
\toprule
Models   & LIME & SHAP \\ \midrule
Random Forests & 0.37  & 0.59      \\
Gradient Boosted Trees & 0.46 & 0.49 \\
Neural Networks & 0.45 & 0.33 \\ \bottomrule
\end{tabular}
\end{table}

Faithfulness (ranging from -1 to 1) reveals the correlation between the importance assigned by the interpretability algorithm and the effect of each attribute on the performance of the model. All of our interpretations receive good faithfulness scores, and SHAP receives a higher faithfulness score than LIME on average. The interpretation by SHAP receives better results because the Shapley value is calculated by removing the effect of specific features which is similar to how faithfulness is computed, so SHAP is more akin to faithfulness.

\begin{table}[H]
\centering
\caption{Monotonicity Evaluation}
\begin{tabular}{@{}ccc@{}}
\toprule
Models   & LIME & SHAP \\ \midrule
Random Forests & False  & False      \\
Gradient Boosted Trees & 22\% True & 22\% True \\
Neural Networks & False & False \\ \bottomrule
\end{tabular}
\end{table}

As for monotonicity, most interpretation methods receive a False though we do find valuable conclusions from interpretations. The difference between faithfulness and monotonicity is that faithfulness incrementally removes each of the attributes, while monotonicity incrementally adds each of the attributes. By incrementally adding each attribute, initially, the model may not be able to make correct predictions with only one or two features, but this does not mean these features are not important. Evaluation metrics for different interpretation methods is still an active research direction, and our results may hopefully stimulate further research on the development of better evaluation metrics for interpreters.


\color{black}

\subsection{\textbf{Summary}}

In this section, the interpretation of four different machine learning models reveals that N-Terminal pro-Brain Natriuretic Peptide (NTproBNP), C-Reaction Protein (CRP), and lactic dehydrogenase (LDH), lymphocyte (LYM) are the four most important biomarkers that indicate the severity level of COVID-19 patients. In the next section, we further validate our methods on two datasets to corroborate our proposal.

\clearpage

\clearpage

\clearpage
\section{Validation on other datasets}

At the initial outbreak of the pandemic, our research leverages a database consisting of patients with confirmed SARS-CoV-2 laboratory tests between 18th January 2020, and 5th March 2020, in Zhuhai, China, and reveals that an increase in NTproBNP, CRP, and LDH, and a decrease in lymphocyte count indicates a higher risk of death. However, the dataset has a limited record of 92 patients which may not be enough to support our proposal. Luckily, and thanks to global cooperation, we do have access to larger datasets. In this section, we further validate our methods on two datasets, one with 485 infected patients in Wuhan, China\cite{Yan2020}, and the other with 5644 confirmed cases from the Hospital Israelita Albert Einstein, at São Paulo, Brazil from Kaggle.

\subsection{\textbf{Validation on 485 infected patients in China}}

The medical record of all patients in this dataset was collected between 10th January and 18th February 2020, within a similar date range as our dataset. Yan et al. construct a dedicated simplified and clinically operable decision model to rank 75 features in this dataset, and the model demonstrates that three key features, lactic dehydrogenase (LDH), lymphocyte (LYM), and high-sensitivity C-reactive protein (hs-CRP) can help to quickly prioritize patients during the pandemic, which is consistent with our interpretation in Table \ref{tab:important_feature}.

Findings from the dedicated model are consistent with current medical knowledge. The increase of hs-CRP reflects a persistent state of inflammation \cite{pmid19411291}. The increase of LDH reflects tissue/cell destruction and is regarded as a common sign of tissue/cell damage, and the decrease of lymphocyte is supported by the results of clinical studies \cite{chen2020epidemiological}.

\begin{figure}[H]
\centering
\includegraphics[width=2.8in]{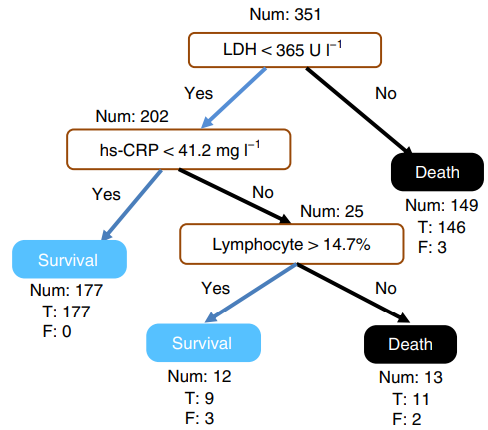}
\caption{A decision rule using three key features and their thresholds in
absolute value. Num, the number of patients in a class; T, the number of
correctly classified; F, the number of misclassified patients. \cite{Yan2020}}
\end{figure}

Our methods reveal the same results without taking efforts to design a dedicated interpretable model but can be more prompt to react to the pandemic. During pandemic outbreak, a prompt reaction that provides insights on the new virus could save lives and time. 

\subsection{\textbf{Validation on 5644 infected patients in Brazil}}

\begin{figure*}
\centering
\begin{subfigure}[b]{0.45\textwidth}
    \centering
    \includegraphics[width=\textwidth]{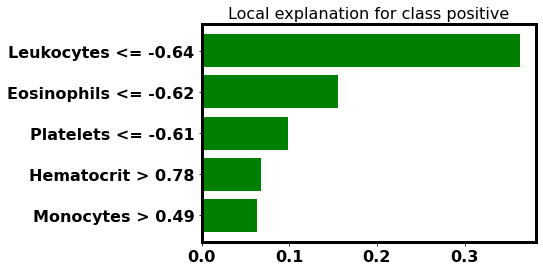}
    \caption{Decision Tree}
    \label{fig:lime_kaggle_dt}
\end{subfigure}
\hfill
\begin{subfigure}[b]{0.45\textwidth}
    \centering
    \includegraphics[width=\textwidth]{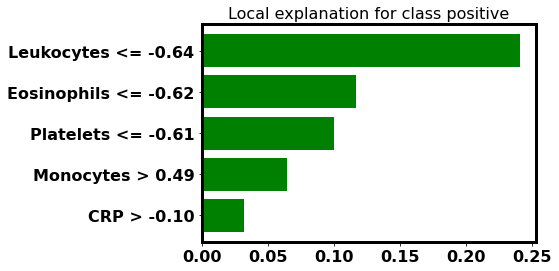}
    \caption{Random Forests}
    \label{fig:lime_kaggle_rf}
\end{subfigure}
\hfill
\begin{subfigure}[b]{0.45\textwidth}
    \centering
    \includegraphics[width=\textwidth]{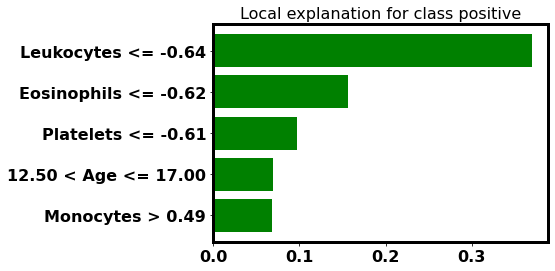}
    \caption{Gradient Boosted Trees}
    \label{fig:lime_kaggle_xgbc}
\end{subfigure}
\hfill
\begin{subfigure}[b]{0.45\textwidth}
    \centering
    \includegraphics[width=\textwidth]{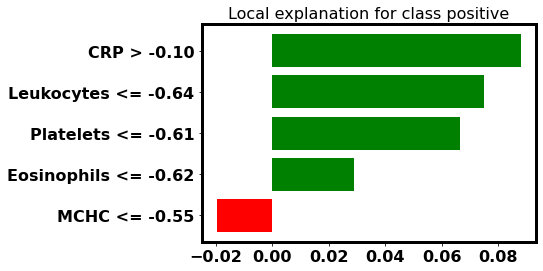}
    \caption{Neural Networks}
    \label{fig:lime_kaggle_nn}
\end{subfigure}
\hfill
\label{fig:kaggle_lime}
\caption{LIME Explanation (\textbf{Kaggle Patient 0}): Features in green have a positive contribution to the prediction (increasing the probability of turning severe), and features in red have a negative effect on the prediction (decreasing the probability of turning severe)}
\end{figure*}

Our approach obtains the same result on the dataset with 92 patients from Zhuhai, China, and a medium-size dataset with 485 patients from Wuhan, China. Besides, we further validate our approach on a larger dataset with 5644 patients in Brazil, from Kaggle.

This dataset consists of 111 features including anonymized personal information, laboratory virus tests, urine tests, venous blood gas analysis, arterial blood gases, blood routine test, among other features. All data were anonymized following the best international practices and recommendations. The difference between this dataset and ours is that all data are standardized to have a mean of zero and a unit standard deviation, thus the original data range that contains clinical meaning is lost. Still, the most important medical indicators can be extracted using interpretation methods.

\begin{table}[H]
\centering
\caption{Patient No.0 in the Kaggle Dataset}
\begin{tabular}{@{}ll@{}}
\toprule
Feature & Value\\ 
\midrule
SARS-Cov-2 test result & 1 \\
Patient Age Quantile & 14.00\\
Hematocrit  & 0.92\\
Platelets  & -1.26\\ 
Mean platelet volume  & 0.79\\ 
Mean corpuscular hemoglobin concentration (MCHC)  & -0.65\\
Leukocytes & -1.47\\
Basophils  & -1.14\\
Eosinophils & -0.83\\
Monocytes & 0.96\\  
Proteina C reativa mg/dL & 0.236\\ 
\bottomrule
\label{tab:kaggle}
\end{tabular}
\end{table}

Following the same approach, a preprocessing is applied on the dataset that removes irrelevant features such as patients' intention to the ward level, and features that have less than 100 patient's record, for instance, urine tests and aerial blood gas tests. On the other hand, patients that have less than 10 records are dropped, because these records do not provide enough information. After preprocessing, we have a full record of 420 patients with 10 features.
\color{black}

\begin{table}[H]
\centering
\caption{Classification Results (Kaggle)}
\begin{tabular}{p{2cm}p{1.1cm}cccp{1cm}}
\toprule
Classifier    & CV & \multicolumn{3}{c}{Test Set} & 95\% confidence interval \\ \midrule
              & F1               & Precision  & Recall  & F1    &\\
Decision Tree & 0.37             & 0.88       & 0.75    & 0.71  & 0.098\\
Random Forests & 0.37             & 0.90       & 0.50    & 0.67  & 0.089\\
Gradient Boosted Trees  & 0.56    & 0.90       & 0.75    & 0.59 & 0.089\\
Neural Networks       & 0.38             & 0.90       & 0.50    & 0.67 & 0.089\\ \bottomrule
\label{tab:kaggle_models}
\end{tabular}
\end{table}

After training and interpreting four different models, decision tree, random forests, gradient boosted trees, and neural networks, the most important features are identified and listed in table \ref{tab:important_feature_kaggle}. The three most common indicative features are leukocytes, eosinophils, and platelets.

According to medical research, patients with increased leukocyte count are more likely to develop critically illness, more likely to admit to an ICU, and have a higher rate of death \cite{zhao2020clinical}. Du et al. noted that at the time of admission, 81\% of the patients had absolute eosinophil counts below the normal range in the medical records of 85 fatal cases of COVID-19\cite{du2020clinical}. Wool G.D. and Miller J.L. discovered that COVID-19 is associated with increased numbers of immature platelets which could be another mechanism for increased clotting events in COVID-19\cite{wool2021impact}.

\begin{table}[H]
\centering
\caption{Five most important features (Kaggle)}
\begin{tabular}{@{}ccc@{}}
\toprule
Model                   & Most Important Features \\ \midrule
Decision Tree           & Leukocytes, Eosinophils, Patient age quantile      \\
Random Forest           & Leukocytes, Eosinophils, Platelets        \\
Gradient Boosted Trees  & Patient age quantile, Hematocrit, Platelets  \\
Neural Networks         & Leukocytes, Platelets, Monocytes          \\  \bottomrule
\label{tab:important_feature_kaggle}
\end{tabular}
\end{table}

In addition, the two datasets collectively reveal that elderly people are more susceptible to the virus. The significant feature NTproBNP in the Chinese dataset is often used to diagnose or rule out heart failure which is more likely to occur in elderly people. And patients that have abnormally low levels of platelets are more likely to be older, male as well \cite{wool2021impact}.
\color{black}

To further validate our interpretation, faithfulness and monotonicity are calculated and listed in tables \ref{kaggle_faith} and \ref{kaggle_mono}. Similarly, our interpretations are consistent with medical knowledge and receive a good faithfulness score, but receive a worse score on monotonicity because the calculation procedure of monotonicity is contrary to faithfulness.

\begin{table}[H]
\centering
\caption{Failthfulness Evaluation (Kaggle)}
\begin{tabular}{@{}ccc@{}}
\toprule
Models   & LIME & SHAP \\ \midrule
Random Forests & 0.71  & 0.82      \\
Gradient Boosted Trees & 0.61 & 0.72 \\
Neural Networks & 0.25 & 0.42 \\ \bottomrule
\label{kaggle_faith}
\end{tabular}
\end{table}

\begin{table}[H]
\centering
\caption{Monotonicity Evaluation (Kaggle)}
\begin{tabular}{@{}ccc@{}}
\toprule
Models   & LIME & SHAP \\ \midrule
Random Forests & False  & False      \\
Gradient Boosted Trees & True & False \\
Neural Networks & False & False \\ \bottomrule
\label{kaggle_mono}
\end{tabular}
\end{table}

\section{Conclusion}

In this paper, through the interpretation of four different machine learning models, we reveal that N-Terminal pro-Brain Natriuretic Peptide (NTproBNP), C-Reaction Protein (CRP), and lactic dehydrogenase (LDH), lymphocyte (LYM) are the four most important biomarkers that indicate the severity level of COVID-19 patients. Our findings are consistent with medical knowledge and recent research that exploits dedicated models. We further validate our methods on a large open dataset from Kaggle and unveil leukocytes, eosinophils, and platelets as three indicative biomarkers for COVID-19.

The pandemic is a race against time. Using interpretable machine learning, medical practitioners can incorporate insights from models with their prior medical knowledge to promptly reveal the most significant indicators in early diagnosis and hopefully win the race in the fight against the pandemic.




%

\clearpage
\appendix[]

\begin{table}[H]
\centering
\caption{Diagnoses}
\begin{tabular}{@{}cc@{}}
\toprule
Feature                   & Comments  \\ 
\toprule
Severity03 & Severe (3) - Normal (0) \\
Severity01 & Severe (1), Normal (0)	\\
\bottomrule
\end{tabular}
\end{table}

\begin{table}[H]
\centering
\caption{Personal Info}
\begin{tabular}{@{}cc@{}}
\toprule
Feature                   & Comments \\ 
\toprule
MedNum           & Medical Number  \\
No            & Patient No.         \\  
Sex  & Man (1), Woman(0)          \\  
Age	& - \\
AgeG1 &	$Age > 50 (1), Age \leq 50 (0)$ \\
Height &  -	 \\
Weight &  - \\
BMI	 & Body Mass Index   \\
\bottomrule
\end{tabular}
\end{table}

\begin{table}[H]
\centering
\caption{Complete Blood Count}
\begin{tabular}{@{}cc@{}}
\toprule
Feature                   & Comments \\ 
\toprule
WBC1 &	White Blood Cell (first time) \\
NEU1 &	Neutrophil Count (first time) \\
LYM1 &	Lymphocyte Count (first time) \\
N2L1 &  -  \\
WBC2 &	White Blood Cell (second time)  \\
NEU2 &	Neutrophil Count (second time) \\
LYM2 &	Lymphocyte Count (second time) \\
N2L2 &  -  \\
\bottomrule
\end{tabular}
\end{table}

\begin{table}[H]
\centering
\caption{Inflammatory Markers}
\begin{tabular}{@{}cc@{}}
\toprule
Feature                   & Comments  \\ 
\toprule
PCT1 &	Procalcitonin (first time)  \\
CRP1 &	C-Reactive Protein (first time)  \\
PCT2 &	Procalcitonin (second time)  \\
CRP2 &	C-Reactive Protein (second time) \\
\bottomrule
\end{tabular}
\end{table}

\begin{table}[H]
\centering
\caption{Biochemical Examination}
\begin{tabular}{@{}cc@{}}
\toprule
Feature                   & Comments  \\ 
\toprule
AST	& Aspartate aminotransferase\\
LDH	& Lactate Dehydrogenase  \\
CK	& Creatine Kinase	 \\
CKMB &  The amount of an isoenzyme of creatine kinase (CK)  \\
HBDH & Alpha-Hydroxybutyrate Dehydrogenase \\
HiCKMB & Highest CKMB  \\
Cr & Serum Creatinine \\
ALB1 &	Albumin Count (first time) \\
ALB2 &	Albumin Count (second time) \\
\bottomrule
\end{tabular}
\end{table}

\newpage
\begin{table}[H]
\centering
\caption{Symptoms and Anamneses}
\begin{tabular}{@{}cc@{}}
\toprule
Feature                   & Comments  \\ 
\toprule
Symptom & -		 \\
Fever & -	\\
Cough & -	 \\
Phlegm & -	 \\
Hemoptysis & -	  \\
SoreThroat & -\\
Catarrh & - \\
Headache & -	 \\
ChestPain & - \\
Fatigue & -	 \\
SoreMuscle & -	 \\
Stomachache & -		 \\
Diarrhea & -	 \\
PoorAppetie & -	 \\
NauseaNVomit & -  \\
Hypertention & -	  \\
Hyperlipedia & -	 \\
DM & Diabetic Mellitus\\
Lung & Lunge Disease	 \\
CAD & Coronary Heart Disease	 \\
Arrythmia & -  \\
Cancer & -  \\
\bottomrule
\end{tabular}
\end{table}

\begin{table}[H]
\centering
\caption{Other test results}
\begin{tabular}{@{}cc@{}}
\toprule
Feature                   & Comments  \\ 
\toprule
Temp & Temperature \\
LVEF & Left Ventricular Ejection Fraction\\
Onset2Admi & Time from onset to admission \\
Onset2CT1 & Time from onset to CT test\\
Onset2CTPositive1 & Time from onset to CT test positive \\
Onset2CTPeak & Time from onset to CT peak \\
cTnITimes &	When was cTnI tested  \\
cTnI & Cardiac Troponin I	 \\
cTnlCKMBOrdinal1  & The value when hospitalized	 \\
cTnlCKMBOrdinal2 & The maximum value when hospitalized \\
CTScore & Peak CT Score  \\
AIVolumneP & Peak Volume \\
SO2 & Empty \\
PO2 & Empty  \\
YHZS & Empty\\
RUL & Empty \\
RML & Empty  \\
RLL & Empty  \\
LUL & Empty  \\
LLL & Empty \\
\bottomrule
\end{tabular}
\end{table}

\clearpage




\ifCLASSOPTIONcaptionsoff
  \newpage
\fi



%



\clearpage
\bibliographystyle{IEEEtran}
\bibliography{IEEEabrv,mybibfile}

%








\end{document}